\documentclass{IEEEcsmag}

\usepackage[colorlinks,urlcolor=blue,linkcolor=blue,citecolor=blue]{hyperref}
\usepackage{tikz}
\usetikzlibrary{shapes, arrows.meta, positioning, fit, backgrounds, calc}

\expandafter\def\expandafter\UrlBreaks\expandafter{\UrlBreaks\do\/\do\*\do\-\do\~\do\'\do\"\do\-}
\usepackage{upmath,color}

\jvol{XX}
\jnum{XX}
\paper{8}
\jmonth{Month}
\jname{Publication Name}
\jtitle{Publication Title}
\pubyear{2021}

\begin{document}

\sptitle{Peer-Reviewed Papers}

\title{Beyond the Black Box:
A Cognitive Architecture for Explainable and Aligned AI}

\author{First Author Hu K.}
\affil{Tongji University, Shanghai, 200000, China}

\markboth{cognitive AI}{Weight-Calculatism}

\begin{abstract}\looseness-1Current AI paradigms, as "architects of experience," face fundamental challenges in explainability and value alignment. This paper introduces "Weight-Calculatism," a novel cognitive architecture grounded in first principles, and demonstrates its potential as a viable pathway toward Artificial General Intelligence (AGI). The architecture deconstructs cognition into indivisible Logical Atoms and two fundamental operations: Pointing and Comparison. Decision-making is formalized through an interpretable Weight-Calculation model (Weight = Benefit × Probability), where all values are traceable to an auditable set of Initial Weights. This atomic decomposition enables radical explainability, intrinsic generality for novel situations, and traceable value alignment. We detail its implementation via a graph-algorithm-based computational engine and a global workspace workflow, supported by a preliminary code implementation and scenario validation. Results indicate that the architecture achieves transparent, human-like reasoning and robust learning in unprecedented scenarios, establishing a practical and theoretical foundation for building trustworthy and aligned AGI.
\end{abstract}

\maketitle

\section{Introduction}

The long-standing synergy between artificial intelligence (AI) and cognitive architectures, particularly Adaptive Control of Thought-Rational (ACT-R) and Predictive Processing(PP), has fostered mutual progress but also led to shared fundamental limitations. However, these paradigms share a fundamental limitation with current mainstream AI: they are architects of experience rather than reasoners of principles, which leads to core challenges in interpretability and value alignment.
Addressing these challenges requires more than incremental improvements to existing models. It demands a foundational re-examination of the computational principles of cognition itself. We propose the "Weight-Calculatism" cognitive theory from first principles, constructing a cognitive model from the ground up at its most fundamental level. Although developed along a different path, it can encompass existing models like the ACT-R and PP architecture, while offering a more profound and generalized framework. Based on the Weight-Calculatism theory, we have the potential to fundamentally resolve the issues of explainability and value alignment.
The distinction can be illustrated by an analogy from chemistry: ACT-R decomposes cognition into functional modules—discrete, functional ‘molecules’ that constitute macroscopic cognitive acts. In contrast, the Weight-Calculatism theory deconstructs these ‘molecules’ into their constituent ‘atoms’ and the ‘forces’ that bind them. It seeks to establish a ‘periodic table’ for cognition by:

reducing all cognitive objects to indivisible Logical Atoms;
decomposing all cognitive processes into two fundamental operations: Pointing and Comparison;
formalizing all decision-making as an on-the-fly computation (Weight = Benefit × Probability), rather than relying on pre-learned weights. This is not a specific ‘molecular’ reaction to be memorized, but a universal principle that drives all ‘cognitive chemical reactions’.

This 'atomic' decomposition confers fundamental advantages over current paradigms:
Radical Explainability: It enables tracing any decision back to its atomic-level constituents and first principles, answering not just how but why.
Intrinsic Generality: It empowers the system to handle novel situations in real-time by composing solutions from atomic operations, rather than relying on a finite library of pre-compiled rules.
Traceable Value Alignment: It grounds the AI's value system in a finite set of auditable Initial Weights, making its goals transparent and its ethical foundation adjustable with precision.
This work represents a shift from macro-functional description to micro-generative principles. Compared with ACT-R, the cognitive process of weight calculation is more transparent, in-depth, and generalizable; compared with PP, the decision-making process of weight calculation is more intuitive, concise, and comprehensive. It lays the foundation for building transparent and value-controllable intelligent systems, and opens up a new path towards robust, trustworthy, and human-compatible general artificial intelligence.\vspace*{-10pt}

\section{The Boundaries of Intelligence: Fundamental Flaws in Current AI Paradigms}

The remarkable achievements of deep learning have cemented the data-driven paradigm as the dominant force in modern AI. However, a closer examination reveals that this paradigm, along with its symbolic predecessors, is approaching a fundamental ceiling. Their shared reliance on extracting and recombining patterns from past experience renders them intrinsically ill-equipped for tasks requiring genuine understanding, robust reasoning, and value-aligned decision-making in novel situations. These theoretical limitations culminate in the practical failures of their most ambitious application: embodied intelligence.\vspace*{-5pt}

\subsection{The Causality Gap in Data-Driven AI}

The core engine of deep learning is correlation, not causation. These models are masters of statistical interpolation within their training distribution, but they lack an internal, manipulable world model. They can associate smoke with fire but cannot represent the causal chain that links them. This leads to well-documented fragility: susceptibility to adversarial examples, catastrophic forgetting, and an inability to generalize systematically beyond their training data. The knowledge they acquire is entangled in billions of uninterpretable parameters, making it opaque, non-modular, and impossible to audit or edit directly—a critical failure for safety-critical applications.\vspace*{-5pt}

\subsection{The Value Grounding Problem in Classical Cognitive Architectures}

Cognitive architectures like ACT-R provide a more structured account of cognition, separating declarative knowledge from procedural skills. Yet, they face a different, equally critical limitation: the value grounding problem. Decision-making in ACT-R is governed by the learned utility of production rules—a metric reflecting past success but lacking foundational justification. The 'why' behind a goal is externally imposed by the modeler, not intrinsically generated by the system.Consequently, while these architectures excel at modelling how a known task is solved, they struggle to explain why a novel goal should be pursued or how to behave when pre-compiled strategies fail. Their intelligence is ultimately retrospective, not prospective.\vspace*{-5pt}

\subsection{A Shared Blind Spot: The Tyranny of Experience and the Marginalization of Logic}

Beneath the apparent dichotomy between connectionist and symbolic approaches lies a profound commonality: both are, at their core, architects of experience. Deep learning models encode statistical experiences from datasets; cognitive architectures encode procedural experiences from task practice. Both are consequently constrained by the past. This shared reliance creates a fundamental inability to tackle genuine novelty—situations that require reasoning from first principles rather than the recombination of past patterns.
Furthermore, both paradigms marginalize explicit logic. In deep learning, reasoning is an emergent property of vector transformations. In production systems, it is embedded in brittle, task-specific rules. Neither possesses a foundational layer of universal logical operations (e.g., for causal relation or comparison) that operates over grounded primitives. This absence is the root of their limited explainability and inability to perform commonsense reasoning reliably.\vspace*{-5pt}

\subsection{The Culminating Failure: Embodied Intelligence and the Infinite Data Trap}

The field of embodied intelligence acts as a crucible that magnifies all the above flaws into a critical failure mode. Placing artificial agents in physical or simulated environments exposes the stark limitations of experience-driven, logic-light paradigms:
The Infinite Data Trap: Learning simple sensorimotor skills (e.g., walking, grasping) often requires millions of trials because the agent is learning statistical mappings without a causal model of its body and world. A slight environmental change can invalidate all learned experience, demanding retraining from scratch.
The Value Vacuum: The goal for an embodied agent is typically an externally imposed reward signal. It learns to walk for a reward, not because it understands mobility is valuable for satisfying fundamental needs (e.g., self-preservation). 
The Commonsense Chasm: A robot might learn to avoid obstacles but does not "understand" why—it lacks the logical atoms and operations to link collision → damage → negation of self-preservation.
These limitations in embodied intelligence collectively demonstrate why experience-driven approaches hit a fundamental ceiling.\vspace*{-5pt}

\subsection{Conclusion: The Case for a Foundational Paradigm Shift}
The limitations discussed—the causality gap, the value grounding problem, the tyranny of experience, and the marginalization of logic—are not mere engineering hurdles. They are symptoms of a deeper philosophical shortcoming: the attempt to construct intelligence without first formalizing its fundamental components—grounded knowledge primitives, universal cognitive operations, and axiomatic values.
The data-driven paradigm and its classical counterparts have taken intelligence as far as possible by recombining experience. The profound challenges in embodied intelligence serve as the ultimate proof that transcending these limits requires a paradigm shift towards cognitive-driven AI, built from the ground up on principles of transparent reasoning and intrinsic value alignment. The following section introduces such a foundation.\vspace*{-5pt}

\section{The Weight-Calculatism Architecture}
Having discussed why a paradigm shift in AI is necessary, this section focuses on the implementation principles of the Weight-Calculative AI Architecture—The Weight-Calculatism cognitive theory, explaining how it resolves these issues. Weight-Calculatism comprises three interlocking components: Logical Atoms, Logical Operations, and the Weight-Calculation Engine.\vspace*{-5pt}

\subsection{Logical Atoms: The Substrate of Cognition}
We posit that intelligence must be built upon stable, interpretable primitives. In Weight-Calculatism theory, Logical Atoms are the fundamental units of cognition. Human cognition ultimately grounds out in intuitive and emotional experiences; what Logical Atoms involve are precisely these conscious experiences and their combinations. Objective entities are not fundamental enough; they are collections of properties, not the most primitive elements of cognition. Here, knowledge-based memories constitute the long-term information repository; episodic memories, i.e., memories of event sequences, essentially involve the ordered storage of several Logical Atoms, which can be names, actions, etc. This constitutes a response to the two types of memory in ACT-R.
For humans, a Logical Atom is a piece of information closely tied to a conscious experience, generated through information input, classification, and storage. Furthermore, information isn't only input from the outside; processing existing information through computation can also generate new information and new concepts, stored as new Logical Atoms. This is the basis for actively expanding and deepening cognition.
Primary, primitive Logical Atoms are generally indivisible. Through processing and integration, primary atoms can evolve to become complex and high-level. The information from a series of Logical Atoms can participate in computation as a whole, which can be termed a higher-level atom. Based on their level of abstraction, they can be roughly divided into three categories:

The most fundamental and concrete are conscious experiences, which originate from perception.
The patterns and regularities of certain conscious experiences—i.e., the common results of computations performed on perceptions—are the properties of things. This is a more abstract category. For example, "universality" is an abstract pattern induced after experiencing phenomena like "Events A, B, and C share the same preconditions, and their outcomes are also the same": it is itself the result of computation on conscious experiences.
Objective entities are collections of properties. This is paradoxically the most abstract and advanced category. For cognition, an objective entity is merely a collection of properties, including concepts crucial to constituting an "object," such as conservation of matter or continuity of motion.

These three categories of basic logical objects originate from perception, computational induction, and integration (broader computation and relation), respectively.
For instance, "position" in our cognition represents the experience, within visual and related perceptions, of an object's form remaining constant while its surrounding environment changes. "Space" represents the totality of positions where objects might exist. "Time" represents the similarities and differences felt between repetitive events (like the alternation of day and night), directly related to the accumulation of memory, and thus possessing continuity. To enable AI understanding, we must construct this entire set of related sensations and patterns in a computationally tractable way, rather than merely representing nouns and their statistical regularities.
Compared to the macroscopic "chunks" in ACT-R, the objects in Weight-Calculatism are more fundamental and concise. To build truly powerful AI, its stored knowledge base cannot remain at the level of nouns like space and time but must deeply construct the relevant patterns and fundamental modes of interaction to achieve genuine understanding.\vspace*{-5pt}

\subsection{Logical Operations: Operating on the Cognitive Substrate}

This section concerns how Logical Atoms interact. Weight-Calculatism theory posits that all logical relations decompose into two fundamental operations: Pointing and Comparison.
Pointing is the activation of one Logical Atom by another. When two Atoms (pieces of information) are strongly related, this operation exists between them. Which Atoms can undergo Pointing needs to be learned and stored beforehand. It embodies relationships: If A, then B. If A and B are events, it can be a "leads to" or "caused by" relationship; if A and B are objects or properties, it can be a "correlates with" relationship, meaning A has property B or property A belongs to B. The "association" in deep learning is implicit, statistical correlation within network weights. In contrast, "Pointing" is an explicit, symbolic link that can itself be inspected and reasoned about.
Comparison involves comparing two streams of information and outputting a "Same" or "Different" result. Identifying crucial differences requires numerous comparison operations on various details. The difference between existence and non-existence (of some qualities) is the foundation of perception and definition; the similarity between two objects is the foundation of induction and analogy. An AI can compare various characteristics of things but doesn't inherently know what these characteristics ultimately signify or what inferences can be drawn.
The objects participating in operations are not only single Logical Atoms but can also be combinations of several atoms, such as "AND" and "OR". For computers, this is implemented by basic logic gates, which need not be discussed in depth here.
The cognitive process is realized by an asynchronous activation propagation algorithm. When a Logical Atom is activated, it triggers all its connected Pointing operations in parallel, propagating activation signals to downstream atoms. Simultaneously, a central working memory collects highly activated atoms and invokes Comparison operations to evaluate them. The entire process lacks a central controller for serial scheduling; the dynamics of cognition emerge from this concurrent, activation-based propagation network.
For example, when the Logical Atom for "smoke" is activated, it deterministically Points to its cause ("fire") and its properties ("toxic"), thereby activating the related atoms. It also Points to its gaseous property and subsequently to the laws governing its next changes ("rising" and "diffusing"). This is not merely the statistical regularity that "smoke is strongly correlated with fire" but the execution of a logical operation. It embodies causality, correlation, and property association in a directly interpretable manner, forming the bedrock of causal reasoning and associative memory.
By feeding the final output information to another processing module that converts it into language or action, a complete chain of learning, thinking, and feedback is formed. The processing and expressive capabilities of the Weight-Calculative AI architecture are more complete, enabling complex symbolic reasoning with a more concise structure. Using deterministic logical relations as the core algorithm is more accurate, efficient, and generalizable compared to vector decomposition and probability calculations.\vspace*{-5pt}

\subsection{Weight-Calculation: The Decision-Making Model}
Logical Atoms and Operations constitute the "thinking" part of the system, while the Weight-Calculation Engine is responsible for "decision-making" and "action." It is an interpretable, human-like decision-making model that unifies rational calculation and emotional drive. The core formula is:
Weight = Benefit × Probability
In the Weight-Calculatism cognitive theory, the actual process is: the brain assigns weights to objects (events, emotions) based on information and fundamental requirements dictated by genes (survival instinct), and then performs calculations. Here, 'Weight' is the result of the computation, representing the priority the brain assigns to an object; it is the significance we assign, not merely its objective value. 'Benefit = Gain - Loss'. Information is processed and computed by the brain to derive the perceived probability of an event occurring, or we can resume P=Connection StrengthonConnection Strengthon+Connection Strengthoff. The comparison of weights corresponding to different potential actions is reflected in consciousness, ultimately determining an individual's thoughts and final behavioral tendencies.
The Weight-Calculation formula describes the human decision-making process, and its objects should be events and (felt) emotions, not physical objects, concepts, or properties. For instance, "money" itself cannot be plugged into this formula, but "obtaining money" can. This event Points to "being able to acquire desired things," which in turn Points to "satisfying material desires"—an initial weight determined by genetic instinct. The specific weight of "obtaining money" is related to its relevance (e.g., the amount of money).
The true power of the Weight-Calculation formula lies in the fact that Benefit can always be decomposed along the causal chain of events into other weights, ultimately tracing back to Initial Weights. Therefore, the weight value of any event must be expressible as:
\begin{equation}
Weight = \sum (Initial Weight_i \times Relevance_i)
\end{equation}
Weight-Calculatism quantitatively incorporates emotion into the weight calculation, positing that emotion itself is generated when the calculation process matches specific patterns. For example:
Acquiring a liked object produces a positive weight, leading to pleasure.
If the actual value exceeds the predicted value, a dopamine-like signal is released. It doesn't directly cause pleasure but enhances motivational strength, driving the agent to act for "acquisition," replicating this unexpected gain.
If an event, left to its natural course, is likely to cause harm to the self, fear is felt, driving the agent to pay attention and make decisions to change the situation.
If this is caused by another agent, anger might also be felt, driving the agent to retaliate to achieve deterrence and prevent future attacks.
The generation of emotion (sensibility) isn't solely about the magnitude of the weight; it involves multiple abstract evaluation systems.
For humans, Initial Weights are determined by genes and instinct. To use the Weight-Calculatism cognitive architecture for building value-aligned decision-making AI, we need only construct a complete, reasonable library of Initial Weights and simulate the patterns that generate emotions. By modifying the Initial Weights, we can easily alter its "personality" and behavioral traits.
The Weight-Calculatism architecture provides a clear, non-anthropomorphic implementation path for affective computing, transforming it from mysterious "qualia" into a designable algorithmic module.
The architecture naturally explains reinforcement learning mechanisms: when the actual benefit resulting from an action significantly exceeds its expected benefit, the system triggers a reinforcement signal analogous to dopamine. This signal does not directly produce pleasure but permanently strengthens the weight or relevance of the "Pointing" chain that led to the successful outcome, making the Weight (W) of that decision path higher in similar future situations. This implements learning from experience and explains the origin of intrinsic motivations such as "the pursuit of surprise" \vspace*{-5pt}.

\section{Implementation Details of Weight-Calculative AI}
The following sections detail the engineering pathways and computational models that bring the Weight-Calculative architecture from theory to practice.

\subsection{Construction Path for Weight-Calculative AI Based on Deep Learning}
To build a new generation of intelligent architecture with deep understanding and value rationality, we propose the following implementation path, based on and transcending deep learning, centered on these core engineering steps:

\subsubsection{Constructing a Fine-Grained Logical Atom Cognitive Library}
The construction of Logical Atoms must follow an "embodied" approach. We need to autonomously learn and distill these stable, interpretable cognitive primitives from multimodal sensor data, building a comprehensive semantic relationship network as the foundation for implementing "Pointing" operations. Manually constructing a basic atom library can accelerate subsequent learning, while the system autonomously learns associative structures from data to continuously enrich this network.

\subsubsection{Implementing Precise Logical Operations}
The "Pointing" operation is embodied as explicit, symbolic links between Logical Atoms, enabling deterministic activation propagation and associative reasoning.
The "Comparison" operation is responsible for decomposing the judgment task of complex information flows into a systematic evaluation of the similarities and differences of basic points. Accurately identifying differences is key to achieving deep understanding and new concept generation and is the core mechanism for constructing new Logical Atoms.
Through these two major operations, we can replace the fuzzy probability predictions in existing systems with clear logical reasoning chains, thereby achieving intelligent behavior that is more efficient and closer to the essence of human cognition.

\subsubsection{Laying a Deep Value Weight Library}
To achieve autonomous decision-making, a value weight library must be constructed by selecting appropriate Logical Atoms (representing fundamental needs or values) and assigning them Initial Weights, forming the axiomatic value foundation for all subsequent Weight-Calculation.
Through the above three steps—constructing fine-grained basic points, implementing precise logical operations, and laying a deep value foundation—we will build a new generation of intelligent architecture with deep understanding and value rationality on top of the perceptual advantages of deep learning.

\subsection{Application of Graph Algorithms: The Computational Engine for Transparent Cognition and Decision-Making}

From a technical implementation perspective, the cognitive library and weight library in Weight-Calculatism theory align highly with the underlying logic of graph algorithms. In fact, we can view the entire Weight-Calculative architecture as a dynamic symbolic computation system operating on a complex graph.
Cognitive Library as a Dynamic Layered Graph:
The most natural representation for the "Logical Atom" cognitive library is a dynamic, layered graph structure. In this graph:
Nodes are logical atoms.
Edges are the various relationships defined by "Pointing" operations.
This graph structure is formally similar to a neural network, but its essence is symbolic and interpretable. Each node and edge has clear semantics.
"Thinking" as an Algorithmic Process on the Graph:
The "thinking" process in the Weight-Calculative architecture can be precisely mapped to specific graph algorithms running on this Logical Atom Graph:
The "Pointing" operation manifests as activation propagation. When a Logical Atom node is activated, this activation acts as a signal that propagates along the node's outgoing edges (i.e., the "Pointing" relationships) to adjacent nodes.
The "Comparison" operation manifests as subgraph matching and alignment. Judging the similarities and differences between two things or situations is essentially performing subgraph similarity computation on the graph. The system uses graph matching algorithms to precisely identify the common nodes ("Same") and differing nodes ("Different") between two subgraph structures.
The fundamental difference from traditional AI graph algorithms lies in the method of graph construction: The graph in Weight-Calculative AI is assembled according to a blueprint by engineers, forming a knowledge structure with explicit semantics; whereas the graph in existing AI (such as knowledge graphs) is more like an original appearance inferred and fitted from data "ruins" by archaeologists, whose completeness and accuracy are limited by the data.

\subsection{Cognitive-Decision Workflow Based on a Global Workspace}

As the cognitive library becomes more complex, the system's demand for information integration and comparison operations grows. Relying solely on a simple linear pipeline is insufficient. The cognitive-decision process is a dynamic cycle coordinated by a Central Computation Area (i.e., the Global Workspace). Functionally corresponding to consciousness in the brain, this hub is the core responsible for integrating information, initiating complex operations, and leading decision-making. It is also key to achieving coherent cognition and transparent explanation within the entire system. Its complete workflow is as follows:

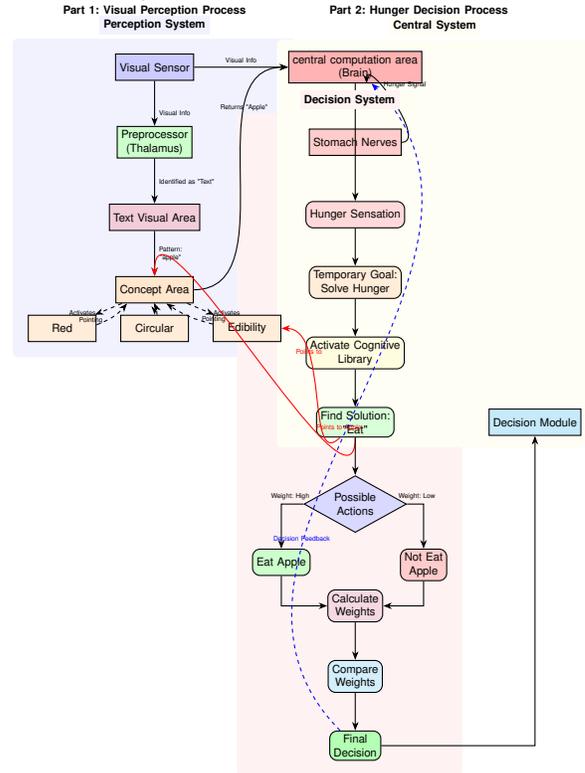
\scalebox{0.5}{
\begin{tikzpicture}[
    node distance=0.6cm and 0.8cm,
    box/.style={rectangle, draw=black, thick, fill=#1, align=center, minimum height=0.7cm, minimum width=1.8cm, font=\small},
    process/.style={rectangle, draw=black, thick, fill=#1, align=center, rounded corners=0.2cm, minimum height=0.7cm, font=\small},
    data/.style={rectangle, draw=black, thick, fill=#1, align=center, dashed, font=\small},
    decision/.style={diamond, draw=black, thick, fill=#1, align=center, aspect=1.8, minimum width=1.2cm, font=\small},
    arrow/.style={-Stealth, thick},
    note/.style={font=\tiny, midway, align=center},
    label/.style={font=\small\bfseries, align=center}
]

\node[label, yshift=0.5cm] (part1) at (-3,0) {Part 1: Visual Perception Process};

\node[box=blue!20] (sensor) at (-3,-1) {Visual Sensor};
\node[box=red!30, right=2.5cm of sensor] (central) {central computation area\\ (Brain)};

\node[box=green!20, below=1.2cm of sensor] (preprocessor) {Preprocessor\\ (Thalamus)};
\node[box=purple!20, below=1.2cm of preprocessor] (visualArea) {Text Visual Area};

\node[box=orange!20, below=1.2cm of visualArea] (conceptArea) {Concept Area};

\node[box=orange!15, below left=0.3cm and 0.5cm of conceptArea] (atom1) {Red};
\node[box=orange!15, below=0.3cm of conceptArea] (atom2) {Circular};
\node[box=orange!15, below right=0.3cm and 0.5cm of conceptArea] (atom3) {Edibility};

\node[label, right=2cm of part1] (part2) {Part 2: Hunger Decision Process};

\node[box=red!20, below=1.2cm of central] (stomach) {Stomach Nerves};
\node[process=red!15, below=1.2cm of stomach] (hungerConscious) {Hunger Sensation};
\node[process=orange!15, below=1cm of hungerConscious] (tempGoal) {Temporary Goal:\\Solve Hunger};
\node[process=yellow!15, below=1cm of tempGoal] (activateLib) {Activate Cognitive\\Library};
\node[process=green!15, below=1cm of activateLib] (findSolution) {Find Solution:\\"Eat"};

\node[box=cyan!20, right=2.5cm of findSolution] (decisionModule) {Decision Module};
\node[decision=blue!15, below=1cm of findSolution] (alternatives) {Possible\\Actions};
\node[process=green!20, below left=0.8cm and 0.5cm of alternatives] (action1) {Eat Apple};
\node[process=red!20, below right=0.8cm and 0.5cm of alternatives] (action2) {Not Eat\\Apple};
\node[process=purple!15, below=1.5cm of alternatives] (weightCalc) {Calculate\\Weights};
\node[process=cyan!15, below=1cm of weightCalc] (compare) {Compare\\Weights};
\node[process=green!30, below=1cm of compare] (finalDecision) {Final\\Decision};

\draw[arrow] (sensor) -- node[note, above] {Visual Info} (central);
\draw[arrow] (sensor) -- node[note, right, pos=0.7] {Visual Info} (preprocessor);
\draw[arrow] (preprocessor) -- node[note, right] {Identified as "Text"} (visualArea);
\draw[arrow] (visualArea) -- node[note, right] {Pattern:\\"apple"} (conceptArea);
\draw[arrow] (conceptArea) to[out=0, in=180] node[note, above, pos=0.7] {Returns "Apple"} (central);

\draw[arrow, dashed] (conceptArea) -- node[note, left, pos=0.8] {Activates} (atom1);
\draw[arrow, dashed] (conceptArea) -- (atom2);
\draw[arrow, dashed] (conceptArea) -- node[note, right, pos=0.8] {Activates} (atom3);

\draw[arrow, dashed, bend right=15] (atom1) to node[note, left, pos=0.3] {Pointing} ($(conceptArea.south west)!0.3!(conceptArea.south)$);
\draw[arrow, dashed, bend right=10] (atom2) to ($(conceptArea.south)!0.5!(conceptArea.south)$);
\draw[arrow, dashed, bend left=10] (atom3) to node[note, right, pos=0.3] {Pointing} ($(conceptArea.south east)!0.7!(conceptArea.south)$);

\draw[arrow] (stomach.east) to[out=0, in=90] node[note, right, pos=0.7] {Hunger Signal} ($(central.south)+(0.3,0)$);
\draw[arrow] (central) to[out=-90, in=90] (hungerConscious);
\draw[arrow] (hungerConscious) -- (tempGoal);
\draw[arrow] (tempGoal) -- (activateLib);
\draw[arrow] (activateLib) -- (findSolution);

\draw[arrow, red, thick] (findSolution) to[out=-135, in=0] node[note, above, pos=0.7] {Points to} (atom3);
\draw[arrow, red, thick] (findSolution) to[out=-90, in=90] node[note, right, pos=0.3] {Points to Apple} (conceptArea);

\draw[arrow] (findSolution) -- (alternatives);
\draw[arrow] (alternatives) -| node[note, above, pos=0.3] {Weight: High} (action1);
\draw[arrow] (alternatives) -| node[note, above, pos=0.3] {Weight: Low} (action2);
\draw[arrow] (action1) |- (weightCalc);
\draw[arrow] (action2) |- (weightCalc);
\draw[arrow] (weightCalc) -- (compare);
\draw[arrow] (compare) -- (finalDecision);
\draw[arrow] (finalDecision) -| (decisionModule);

\draw[arrow, dashed, blue, thick] (finalDecision) to[out=135, in=-45] node[note, above, pos=0.3] {Decision Feedback} (central);

\begin{scope}[on background layer]
    \node[fill=blue!5, rounded corners, fit=(sensor) (preprocessor) (visualArea) (conceptArea) (atom1) (atom2) (atom3), inner sep=0.4cm] (perceptionArea) {};
    
    \node[fill=red!5, rounded corners, fit=(stomach) (hungerConscious) (tempGoal) (activateLib) (findSolution) (alternatives) (action1) (action2) (weightCalc) (compare) (finalDecision), inner sep=0.4cm] (decisionArea) {};
    
    \node[fill=yellow!5, rounded corners, fit=(central) (decisionModule), inner sep=0.3cm] (centralArea) {};
\end{scope}

\node[fill=blue!5, font=\small\bfseries, above=0.1cm of perceptionArea] {Perception System};
\node[fill=red!5, font=\small\bfseries, above=0.1cm of decisionArea] {Decision System};
\node[fill=yellow!5, font=\small\bfseries, above=0.1cm of centralArea] {Central System};

\end{tikzpicture}
}

\subsubsection{Phase 1: Multimodal Perception and Preprocessing Activation}
Receptor Signal Input: External information acquired by receptors (e.g., visual, somatosensory) is sent in parallel to the Central Computation Area (Global Workspace) and the corresponding Preprocessors.
Preprocessing and Pattern Recognition: The preprocessors (functionally analogous to the thalamus and its associated specific sensory cortices) perform preliminary classification and pattern recognition on the raw signals. For example, the visual preprocessor identifies the input as a "text pattern," routes it to the "text visual area" for fine decoding, and finally outputs the recognition result, such as the string "apple".
Initial Activation Injection: The output from the preprocessors is sent to the Central Computation Area. At this point, the Central Computation Area, acting as an integration and scheduling center, begins to dominate the subsequent cognitive processes.

\subsubsection{Phase 2: Centrally-Driven Concept Activation and Situation Building}

Central Initiation of Concept Retrieval: Upon receiving preprocessed information (e.g., "apple"), the Central Computation Area does not wait passively but actively issues retrieval commands to the conceptual storage area (the declarative memory, where "apple" corresponds to a third-level, or conceptual-level, Logical Atom).
Logical Atom Activation and "Pointing" Operation: The corresponding Logical Atom (i.e., the "apple" concept node) in the storage area is activated. This atom, through its built-in "Pointing" operation, automatically evokes a cluster of strongly associated property atoms (e.g., Level 1: "red," "round," "sweet"; Level 2: "edibility," "fruit category").
Information Feedback and Situation Integration: These activated property pieces of information are fed back to the Central Computation Area. The hub, at this moment, binds and integrates information from different preprocessing channels (e.g., the visual shape of the "apple," the somatosensory "hunger" signal) to form a unified, multi-dimensional Current Situation Understanding Graph within the workspace. This corresponds to consciousness integrating a vague perception into a meaningful "scene."

\subsubsection{Phase 3: Goal-Oriented Weight-Calculation and Decision Arbitration}

Goal Generation and Setting: Based on internal states (e.g., the received "hunger" signal) and the Initial Weights, the Central Computation Area dynamically generates or sets the current goal (e.g., "resolve hunger").
Central Initiation of Solution Search: Using this goal as a new starting point, the hub again actively commands the cognitive library to perform a search, traversing possible action paths (e.g., "eat") via "Pointing" operations.
Weight-Calculation and Option Evaluation: For each candidate path (e.g., "eat the apple," "do not eat the apple"), the Central Computation Area coordinates the Weight-Calculation Engine to simulate their potential consequences and calculate the Weight for each consequence (Weight = Benefit × Probability). For example, it calculates the satiety (positive benefit) from "eating the apple" against the consumption of food reserves (negative benefit) and their respective Relevance (how much satiety is increased; how much value of food is consumed).

Decision Arbitration: The Central Computation Area collects the final Weight-Calculation results for all candidate paths and, acting as the final arbitrator, compares and selects the action with the highest calculated Weight for execution. The entire calculation process is fully transparent to the hub, laying the foundation for generating explanations.

\subsubsection{Phase 4: Action Execution and Experience Consolidation}
Motor Command Issuance: The Central Computation Area translates the decision into specific motor commands. For skilled actions (e.g., "pick up the apple and chew"), this command may directly invoke pre-stored programmed motor sequences for efficient execution.
Learning and Graph Update: The outcome of the action serves as a feedback signal. The "Pointing" connections on successful action paths are strengthened; entirely new experiences are integrated into the cognitive library, forming new Logical Atoms and relationships, thereby enriching the system's cognitive graph.

\subsection{Practicalization and Scalability: Building a Portable, Evolvable General Intelligence Foundation}
The revolutionary potential of the Weight-Calculative architecture lies not only in its cognitive depth but also in its engineering characteristics of scalability and practicalizability. Its core—the Logical Atom cognitive library and the Initial Weight value library—serves as a decoupled, symbolic knowledge-value system, laying a solid foundation for achieving this goal.

\subsubsection{Portability and Domain Adaptation of Core Libraries}

Unlike deep learning models with highly entangled and difficult-to-interpret parameters, the core assets of Weight-Calculative AI are highly structured.
Cognitive Library as Knowledge Graph: The Logical Atom Graph is essentially a highly expressive, portable knowledge graph. A basic cognitive library pre-trained and constructed in a general domain can serve as a "world model" seed and be quickly transplanted to specific domains (e.g., healthcare, finance, autonomous driving).
Ethical Fine-Tuning of Value Library: The Initial Weight library is also portable. We can load the AI with a widely validated "basic moral chassis" aligned with human interests. When deploying to a specific domain, instead of retraining the entire model, experts can finely adjust (Fine-tuning) the specific relevance strengths of a small number of initial weights to achieve safe and reliable value alignment. 

\subsubsection{Redundancy and Openness: The Key to Handling Long-Tail Problems}
The fragility of current AI largely stems from the closed nature of their training data and the over-optimization of their models. The Weight-Calculative architecture addresses this by embracing functional redundancy and openness.
Redundancy Ensures Robustness: Just as the human brain possesses cognitive associations far exceeding daily needs, a large-scale Logical Atom cognitive library naturally contains a vast amount of "redundant" information. For example, for "abnormal road surface reflection," the system can understand it not only from the "driving" perspective but also potentially invoke logical atoms from multiple dimensions like "physics - specular reflection," "meteorology - water accumulation," or even "art - luster" for cross-validation and reasoning. This multi-path knowledge access mechanism is the fundamental reason it can handle long-tail cases not seen during training (e.g., being misled by oil stain reflection on the road).
Continuous Evolution: The system can continuously learn, constantly introducing new logical atoms and "Pointing" relationships to enrich the cognitive library, without facing "catastrophic forgetting" like neural networks. This builds an agent capable of lifelong learning and continuous evolution.

\subsubsection{Hybrid Decision-Making Mode: Balancing Efficiency and Innovation}
In practical applications, pure serial symbolic reasoning may not meet the real-time requirements of all scenarios. Therefore, we propose a hybrid decision-making mode:
Reflexive Fast Channel: For high-frequency, high-risk decisions with clear patterns (e.g., emergency braking), the system can solidify highly successful "Pointing" paths to form near-instinctive reflex arcs, achieving millisecond-level response.
Deliberative Slow Channel: For novel, complex decisions or those involving significant value trade-offs (e.g., strategy selection for an unprotected left turn), the central computation area initiates the full Weight-Calculation reasoning process for innovative thinking and prudent planning.
This "dual-channel" design (fast and slow) allows Weight-Calculative AI to handle routine tasks as efficiently as traditional AI, while also demonstrating common-sense reasoning and value judgment capabilities beyond the data scope at critical moments.

\subsubsection{Modular Architecture and Central Scheduling}
To achieve scalable application, Weight-Calculative AI should adopt a highly modular architectural design:
Functional Modules: The system contains multiple dedicated modules, such as visual recognition, language processing, motor control, etc. Each module is responsible for preliminary signal processing and symbolization locally, providing the central computation area with structured streams of logical atoms.
Core Role of the Central Computation Area: The central computation area acts as the system's "CEO". It does not process raw data directly but is responsible for high-level scheduling and decision-making. It receives inputs from various modules, performs information fusion and deep reasoning in the Global Workspace, invokes the Weight-Calculation engine to make decisions, and finally distributes abstract action commands to the corresponding execution modules.
This architecture ensures clear division of labor, ease of extension, and maintenance, while also allowing the central hub to focus on core cognitive and decision-making tasks, providing a feasible engineering path for building complex, large-scale multimodal intelligent systems.

Summary: Through the portability of core libraries, tolerance for redundancy and openness, hybrid decision-making modes, and modular architecture, Weight-Calculative AI demonstrates a clear path towards scalability and practicalization. It aims to build not just a tool for solving single tasks, but a general intelligence foundation that can be deployed across domains, continuously learn and evolve, with transparent behavior deeply aligned with human values.

\section{Feasibility Analysis and Validation}

To validate the feasibility of the Weight-Calculatism architecture, we designed two progressive experimental scenarios: a classic moral dilemma to test decision transparency and value traceability, and an unprecedented encounter with an alien ecosystem to challenge its analogical reasoning and dynamic learning capabilities in completely novel situations.

\subsection{Core Model Validation: Transparent Decision-Making in Moral Dilemmas
}We constructed a computational model of a fire escape scenario where a hungry scholar must choose between saving scientific notes crucial for civilization or saving canned food for personal survival.(The complete specification of the Logical Atom cognitive library, including the detailed taxonomy and relationship types, is provided in the Supplementary Materials, available in the IEEE Computer Society Digital Library.)
The system initializes a cognitive library of Logical Atoms connected via Pointing operations into a knowledge network. The value system is defined by a refined weight library containing a finite set of meaningful Initial Weights.
Upon receiving perceptual input, activation spreads through the network. The decision engine then calculates the total weight for each candidate action using the formula $Weight = \sum (Initial Weight_i \times Relevance_i)$, where Benefit traces back to Initial Weights and Probability encodes relationship strengths.
Results showed that "carry scientific-notes" achieved a total weight of 20.80, superior to "carry canned-food" (0.80). Crucially, the system provided a human-readable explanation:
Radical Explainability: It traced and displayed every reasoning path, transparently quantifying the high positive weight from contributing to civilization continuation and the negative weight from increased death risk.
Traceable Value Alignment: All weights were directly linked to the finite, auditable set of Initial Weights.
Transparent Arbitration: The final decision was based solely on the explicit weight calculation formula.
This experiment confirms that the Weight-Calculatism architecture achieves human-level complex trade-offs and provides the deep explanation rare in machine decision-making. 

\subsection{Novel Response Validation: Analogy and Dynamic Learning in an Alien Ecosystem}
An alien ecosystem risk assessment scenario was designed to test the architecture's ability to handle genuinely novel situations beyond its training data. The AI encountered alien lifeforms with unknown features and had to decide between "immediate research," "cautious retreat," and "remote monitoring."
Using Earth biology as a cognitive baseline, the system performed Comparison operations across structural, functional, and behavioral dimensions. It identified a high novelty score, with an overall similarity to Earth biology of only 0.106.
Confronted with the unknown, the system engaged in dynamic learning. Based on partial similarities, it dynamically created new learned relations, simulating human scientists' ability to form hypotheses by analogy.
Using the expanded cognitive graph, the system re-evaluated action weights. "remote-monitoring" was selected as the optimal decision, balancing the opportunity for scientific discovery against crew safety risk. The system explained: "Novel biological features detected with significant differences from Earth lifeforms... Balanced approach ensures crew safety while maintaining scientific observation."
This validates the architecture's core strength: innovatively responding to completely novel situations through first-principles reasoning and dynamic knowledge expansion, rather than relying on pre-existing pattern matching—a key capability for AGI.

\subsection{Engineering Feasibility of Complex Weight-Calculative AI}
The experiments validate the architecture's principles. Scaling it into a complex system requires a clear engineering path addressing core challenges.
\subsubsection{Feasible Path for Cognitive Library Construction}
Building a comprehensive Logical Atom library is a large-scale endeavor but achievable through layered, automated, and incremental strategies.
Layered Strategy:
Core Layer: Manually refine 100-500 fundamental atoms (e.g., space, time, causality, force) as the cognitive foundation.
Domain Layer: Modular expansion into specialized fields (e.g., physics, biology, society), using existing knowledge graphs (e.g., ConceptNet) as initial seeds.
Instance Layer: Automate the population of specific objects and properties using LLMs for information extraction from text.
Automation Support: Guided by the core layer, LLMs and reinforcement learning from environmental interaction can semi-automate the establishment and verification of "Pointing" relations.

\subsubsection{Safe Construction of the Weight Library and Value Alignment}
The Weight Library, though small, is the cornerstone of value alignment and must be built with the highest safety standards.
Minimal Privilege Principle: The set of Initial Weights must be minimal, fundamental, and universal (e.g., "survival," "knowledge," "avoidance of harm").
Auditability: Each Initial Weight requires clear philosophical and ethical justification.
Societal Calibration: Final determination of the weight set should involve interdisciplinary expert committees and broad public participation to avoid cultural bias.

\subsubsection{Scalable Technical Architecture}
A layered and partitioned technical architecture is designed to address the computational complexity of large-scale cognitive graphs.
Vectorized Graph Representation: Embedding vectors for each Logical Atom enable efficient semantic similarity calculations alongside precise symbolic reasoning.
Partitioned Architecture: The global cognitive graph is naturally partitioned into sparsely connected sub-graphs by domain, allowing most reasoning to occur within sub-graphs, drastically reducing computational load.
Central Scheduling Model: The Global Workspace acts as the system's "CEO," sending queries to cognitive partitions and integrating summary results, enabling scalable complex decision-making.
Hybrid Decision-Making: Combines a fast Reflex Channel (using solidified "Pointing" paths for high-frequency decisions) with a slow Deliberative Channel (performing full Weight-Calculation for novel scenarios), balancing real-time response with deep reasoning.
Existing technologies like distributed graph databases, graph neural network frameworks, and high-performance computing provide a solid foundation for implementing this architecture. Therefore, building a practical Weight-Calculative AI system, while engineeringly challenging, is a feasible path with clear principles and a defined route.

\section{CONCLUSION}
The development of Weight-Calculative AI is not only necessary and feasible, but although it remains at a conceptual stage, we now possess a clear, actionable path to build it step by step.

\section{ACKNOWLEDGMENTS}

The authors are grateful to the anonymous reviewers for their insightful comments. In the preparation of this work, the authors used AI-assisted tools for translation support, code implementation, and literature research. After using these tools, the authors reviewed and edited the content as needed and take full responsibility for the content of the publication.

\def\refname{REFERENCES}

\begin{IEEEbiography}{Hu Keyi}{\,} is a student in artificial intelligence at Tongji University. His research focuses on cognitive architectures, explainable AI, and value alignment. Contact him at HuKey@tongji.edu.cn.
\end{IEEEbiography}

\end{document}